\documentclass{article}

\usepackage{arxiv}

\usepackage[utf8]{inputenc} 
\usepackage[T1]{fontenc}    
\usepackage{hyperref}       
\usepackage{url}            
\usepackage{booktabs}       
\usepackage{amsfonts}       
\usepackage{nicefrac}       
\usepackage{microtype}      
\usepackage{lipsum}
\usepackage{graphicx}
\graphicspath{ {./images/} }

\usepackage{amsmath, amssymb, bm}
\usepackage{ color}
\usepackage{float}
\usepackage[linesnumbered,ruled,vlined]{algorithm2e}

\usepackage{bigints} 

\newcommand{\ind}{\mbox{$\perp\!\!\!\!\!\perp$}}

\title{Enhancing Variable Importance in Random Forests: A Novel Application of Global Sensitivity Analysis}

\author{
 Giulia Vannucci \\
  Department of Electrical Engineering and Information Technology\\
  Polytechnic and Basic Sciences School\\
  University of Naples Federico II\\
  Via Claudio, 21\\
  80125 Napoli (NA), Italy
  \texttt{giulia.vannucci@unina.it} \\
   \And
 Roberta Siciliano \\
  Department of Electrical Engineering and Information Technology\\
  Polytechnic and Basic Sciences School\\
  University of Naples Federico II\\
  Via Claudio, 21\\
  80125 Napoli (NA), Italy
  \texttt{giulia.vannucci@unina.it} \\
  \And
Andrea Saltelli \\
 University Pompeu Fabra \\
       Barcelona School of Management\\
       Carrer de Balmes, 132\\
       08008, Barcelona, Spain\\
Centre for the Study of the Sciences and the Humanities\\
University of Bergen\\
Parkveien 9, PB 7805\\
5020, Bergen, Norway
  \texttt{andrea.saltelli@gmail.com}
}

\begin{document}
\maketitle
\begin{abstract}
The present work provides an application of Global Sensitivity Analysis to supervised machine learning methods such as Random Forests. These methods act as black boxes, selecting features in high--dimensional data sets as to provide accurate classifiers in terms of prediction when new data are fed into the system. In supervised machine learning, predictors are generally ranked by importance based on their contribution to the final prediction. Global Sensitivity Analysis is primarily used in mathematical modelling to investigate the effect of the uncertainties of the input variables on the output. We apply it here as a novel way to rank the input features by their importance to the explainability of the data generating process, shedding light on how the response is determined by the dependence structure of its predictors. A simulation study shows that our proposal can be used to explore what advances can be achieved either in terms of efficiency, explanatory ability, or simply by way of confirming existing results.
\end{abstract}


\section{Introduction}

Machine learning (ML) techniques are increasingly used in a number of scientific domains, such as engineering and life sciences, where ML is already an established tool, and behavioural social sciences, where ML is currently spreading. Among the variety of ML algorithms, tree--based methods are a popular class of predictive models because they are conceptually simple, powerful in both prediction and classification tasks, and able to deal intrinsically with non-linearities and interactions \cite{hastie2009}. The famous CART algorithm \cite{breiman1984} is the basis of the most widely used ensemble algorithms such as Random Forests \cite{breiman2001a}. Note that when moving from a single tree--based model to an ensemble model, the loss of interpretability of the model becomes a problem whenever one is interested in understanding the relationships between features. In fact, ensemble methods are usually regarded as black boxes, very useful for making predictions but less able to provide insight into how the method arrived at its prediction \cite{rothacher2023}. A key tool of tree--based algorithms is the Variable Importance (VI) measure, which captures the magnitude of the contribution of each predictor to the response in terms of the final prediction. However, there are several examples where variable importance is used to gain insight into the dependence between the response and the explanatory variables in a generative sense \cite{hu2020, dhingra2023}. The clear distinction in the literature between the predictive and explanatory purposes of statistical modelling \cite{breiman2001b, cox2001, shmueli2010} raises the question of whether an ML model that is predictive by design can also provide some information about the data generation process. 

In the context of ML, interpretability means understanding why a particular prediction is made so that decisions can be made, knowing which input features drive the predictions, and making sense of the prediction itself \cite{iooss2022}. There is a lack of clarity in the literature about rigorous definitions of the concepts of interpretability and explainability: although they are two related terms, they are not identical, as these concepts refer to deep cognitive processes related to the social sciences and their different fields of application or scientific communities \cite{iooss2022}. Moreover, a rigorous definition of these concepts is required by the legislation on data processing: In 2018, with the introduction of the General Data Protection Regulation (GDPR) by the European Parliament, which for the first time introduces, to some extent, a right to explanation for all individuals to obtain meaningful explanations of the logic involved when automated decisions are taken \cite{guidotti2018}, which is linked to concepts such as {\it fairness} and {\it transparency}\footnote{A ML model is transparent {\it if the fit of the ML model to training data and its generalisation to fresh data can be described and motivated by the analyst with fair documentation that allows its reproducibility}. }.

In this paper we distinguish between two concepts. We refer to the {\it explanatory ability} when we focus on understanding which predictors and how they interact to have a direct effect or influence on response variation. {\it Interpretability}, on the other hand, focuses on understanding the inner workings of the ML model. A linear regression model provides parameter estimates and p-values, a single tree provides a connected and oriented graph. These ML models are more interpretable than ensemble methods such as random forests. 

{\it Explainability} focuses on explaining the decision or prediction result and refers to the ability to justify the results of a ML model. Explainability in Explainable ML focuses on making the complex decisions and outputs of a ML model understandable to users, regardless of their technical expertise. 
Explainable ML includes transparency, interpretability and explanatory ability. 

Global Sensitivity Analysis (GSA) explores the relative influence of uncertain input factors in determining the uncertainty in the model prediction \cite{saltelli2002a}. A brief history of the discipline can be found in \cite{tarantola2024}, while recent reviews can be found in \cite{razavi2021, saltelli2021}. GSA has been applied to the study of variable selection in regression \cite{becker2021}, suggesting a similar, though different, application to feature selection in ML. Whereas in \cite{becker2021} the emphasis was on finding a more efficient way of identifying a known data generating process, here the emphasis is on the interpretability of the results. Note also that when performing a GSA, the question is not only {\it which factors are influential}, but also {\it how they are influential}, for example whether interactions are relevant and what is the effective dimension of the problem \cite{Puy2022}. GSA can be usefully applied to identify non--influential factors and thus assist in model simplification \cite{Saltelli2002b}. Considerable analogies have been identified in the problem setting of GSA and ML \cite{razavi2021, iooss2022}: whereas in ML one seeks a function that can map from an input space to an output of interest, generally independent of any underlying theory, physical or otherwise (sometimes the function acts as a classifier for the output), in GSA one does not seek to identify a specific function, but to find the relative strengths of input--output relationships via coefficients such as Sobol' indices or Shapley effects. 
    
As noted in \cite{razavi2021}, there are differences between these two settings. In most cases, computational experiments, for example based on a system of differential equations, are deterministic; this is not the case in ML, where observational data come with an error. Again, computational experiments generally require a theory, whereas the effectiveness of ML is often linked to its independence from an existing theory. ML requires large input data, whereas numerical experiments can and often are performed in situations where data are scarce. Finally, when using mathematical models in numerical experiments, one can design the experiment (where to select data points from, for example using a fractional factorial or quasi--random design), whereas this is not the case in ML.

Despite these differences, there are several experiments where GSA is used in conjunction with ML. It has been shown that when analysing the importance of features in ML, the so--called total sensitivity indices \cite{homma1996} described below are close to the permutation--based importance measures traditionally used in ML \cite{antoniadis2021}. The last reference not only shows the exact relationship between the overall sensitivity index and the permutation--based feature importance, but also demonstrates the use of ML \textit{prior} to GSA as an efficient way to prune the number of variables before computing the sensitivity indices. 

Can the GSA approach identify a variable importance ranking for understanding the true data generating process underlying the data?

In this paper, we aim to answer the question: 
{\it ``Can the GSA approach construct a variable importance measure that can help in understanding the true data generating process underlying the data?''}. 
Rather than proposing a new algorithm that can outperform the alternatives, we want to propose a combined approach in which our proposal can give the {\it generative importance rankings of VI} with respect to other methods of calculating the {\it predictive VI}. As we will see, the discrepancy between the two interpretations can be informative for the analyst. This places our proposal in the context of the branch of {\it ``Explainable ML''} that involves the development of strategies to penetrate the ``black boxes''. Specifically, we show the performance on ranking with VI computed with our approach, namely Random Forests Global Sensitivity--based Variable Importance, and with: the classical measures proposed for CART \cite{breiman1984} and RF \cite{breiman2001a}; the conditional VI proposed by \cite{strobl2008} for conditional inference RF \cite{hothorn2006}; the Sobol MDA measure proposed by \cite{benard2022}. 

The paper is structured as follows: Section \ref{sec:rtrf} gives a brief overview of RT and RF methods. Section \ref{sec:gsa} introduces GSA, with a review of sensitivity measures in Section \ref{sec:sm} and the general discussion of the GSA paradigm in Section \ref{sec:gsa-par}. Our proposal is described in Section \ref{sec:GS-VI}. The results of a simulation study comparing our proposal with others are discussed in Section \ref{sec:sim}, while two applications are presented in Section \ref{sec:appl}. Some conclusions are drawn in the last section.

\section{Tree--based methods for Regression}\label{sec:rtrf}

Tree--based methods are used in many fields of applications (medicine, genetics, finance, marketing, etc.) as a supervised machine learning model for complex data structure. They follow a non--parametric distribution free approach to deal with non--linearity in the relationship between a set of explanatory variables or predictors and a target or response variable. Classification and Regression Trees (CART) is the benchmark methodology \cite{breiman1984}.

Focus of this paper is the regression problem. Let $(Y, {\bf X})$ be a multivariate random variable where ${\bf X}$ is the vector of $K$ features playing the role of input or explanatory variables taking values in $\Xi \in {\cal R}^K$ and $Y$ is the numerical output or response taking value in the real space ${\cal R}$. 

A supervised learning model for regression aims to identify a {\it classifier or predictor} $d({\bf x})$ on the basis of a learning sample ${\cal L} = \{(y_n, {\bf x}_n), n = 1, \dots N\}$ taken from the multivariate distribution of $(Y, {\bf X})$ such that for every measurement ${\bf x}$, $d({\bf x})$ is equal to a value in ${\cal R}$. 

\subsection{Regression Trees}

Regression Tree (RT) $T(\bm{X})$ is a tree structured predictor that is identified starting from the binary partition of $\mathcal{X}$ at the root node into disjoint subsets and proceeding recursively for each of them up to the final partition of $\mathcal{X}$ into a set of terminal nodes or leaves where is assigned a constant response value. At each internal node, the best split is found on the basis of the binary partition of one of the predictors according to a goodness of split criterion. Regression tree growing attempts to reduce recursively the within--node variability of the response values, finally also in the leaves.

The RT model can be formalized as

\begin{equation}\label{eqtree}
T(\bm{X})=T(\bm{X}; \mathcal{R}_Y, \bm \mu)=\sum_{m=1}^M \mu_{R_m} \mathbb{I}_{(\bm X \in R_m)}
\end{equation} 
where $\mathcal{R}_Y=(R_1,\dots,R_M)$ is a binary recursive partition of $\mathcal{X}$ :  $\mathcal{X} = \bigcup_{m=1}^{M}R_m$, ${\mu}_{R_m}=\mathbb{E}(Y | \bm {X}\in R_m)$, the means of $Y$ within the terminal regions (or nodes), are the values that minimize the MSE of the tree, and $\mathbb{I}_{(\bm {X} \in R_m)}$ is the indicator variable of the elements within the partition. The construction of the tree moves around three elements: the splitting criterion, that refers to the choice of the split variable $X_j$ at the split point $s$ such to decrease the Mean Squared Error (MSE) of the tree $MSE=\mathbb{E}(Y- \hat{T}(\bm{X}))^2$; the stopping rule to declare a node to be terminal such as the minimum value of observations (i.e. $5$); the prediction rule that labels each leave by the average of the response values within that leave. A tree path can be understood as a production rule describing the interaction among the features yielding to a response value. Any tree structure $T(\bm{X})$ with a fixed number of leaves can be very useful for exploratory purpose to understand and visualize the dependence data structure, specifically the interaction among the features that better explain the variability reduction. 

As it concerns the tree predictor $d({\bf x})$, the generalization mean squared error can be decomposed in the sum of the variance and the squared bias. An induction strategy is required to identify the honest size tree that can be generalized for fresh data. This requires compromising two types of errors in the famous well known \emph{trade--off bias versus variance}: a too large tree might overfit the training data (i.e., high variance) whereas a too small tree may not capture the data generating process (i.e., high bias). Tree size is the tuning parameter governing the complexity of the model, i.e. the VC--dimensionality, \cite{vapnik1989}, and the optimal tree size should be adaptively chosen from the data using an independent test set or cross--validation (CV). 

The induction procedure provided by CART is in two steps: cost--complexity pruning procedure and decision tree selection. The first aims to identify a sequence of nested subtrees $T_{max} \subset T_1 \dots \subset {t_1}$ where $\{t_1\}$ is the root node of the tree and $T_{max}$ is the maximum expanded tree, with the associated sequence of tree sizes from the highest to the lowest. The final decision tree is then selected minimizing the MSE for regression using an independent test sample. Alternatively, a resampling procedure such as a $k$--fold CV estimate can be preferred. A one--step visual pruning and tree selection has been recently introduced \cite{iorio2019}. 

\subsection{Random Forests}

More effective for prediction are multiple classifiers or ensemble methods \cite{dietterich2000}. These consist in the construction of a set of classifiers (weak learners) by re--sampling the data of the training sample and then in classifying new data points by averaging their tree predictions. A necessary and sufficient condition for an ensemble of classifiers to work better than any single classifier is that the classifiers to be aggregated must be accurate and diverse. 
Bagging \cite{breiman1996} uses bootstrap replication, whereas Boosting \cite{freund1999} uses a weighted--bootstrap replication at each iteration such to force the algorithm to learn by its errors becoming a strong learner. 

Random Forests (RF) \cite{breiman2001a} provide an improvement over bagging. It is an ensemble method that combines several individual regression trees such that each tree depends on a random vector of units sampled independently and with the same distribution for all trees in the forest. The algorithm of RF begins with sampling with replacement $n_h$ units from the original learning set: only the sampled units are employed for a tree construction. Each random sample reflects the same data generating process, but differs slightly from the original training sample because of random variation. Then, the growth of $H$ trees is done as in CART, with the peculiarity that only a subset of dimension $l<p$ of variables is considered at each node of the construction of the tree. The random forests predictor is 

\begin{equation}\label{rfest}
T_{\texttt{rf}}(\bm {X})=T(\bm {X}_{h}^{*}; \mathcal{R}_{Y_{h,n^{*}}}, \mu_{h,n^{*}} )=\frac{1}{H}\sum_{h=1}^H T (\bm{X}_{h}^{*}; \mathcal{R}_{Y_{h,n^{*}}}, \bm \mu_{h,n^{*}}))
\end{equation}
where $H$ is the total number of trees in the forest, $\bm{X}_{h}^{*}$ denotes the random subset of explanatory variables sampled for each node of the $h^{th}$ tree in the ensemble, $\mathcal{R}_{Y_{h}}$ and  $\bm \mu_{h}$ are respectively the regions and the mean on the bootstrapped sample that defines the  $h^{th}$ tree. Therefore, the number of randomly preselected splitting variables and the total number of trees in the forest are parameters that affect the stability of the results of RF.

An important feature of this methodology is their use of out--of--bag samples, the bootstrapped data which are not used to fit the trees, for the out--of--bag error estimates. The Out--Of--Bag estimate (OOB) is calculated by predicting the real value for each observation in the train set $({X}_i, Y_i)^{\mathtt{train}}$ by using only the trees for which this observation was not included in the bootstrap sample. As proved by \cite{breiman1996}, the OOB estimates are as accurate as using a test set with size equal to the training set. Therefore the use of the OOB estimates removes the need of splitting the data in training and test set. 

\subsection{Variable Importance}

RTs and RF can be used to rank the predictive importance of the variables. For RTs, \cite{breiman1984} defined the VI measure that reflects the relative importance, or contribution, of each input variable in predicting the response. If it is simple to think about the contribution of a splitting variable like the relative improvement in the deviance of the model, more difficult can be ranking those variables that never occur in the tree structure. With this aim, the VI can be computed with the support of the surrogate split. 

How variables are ranked can be critical for those that do not return the best split, but the second or third best split. For example, it may happen that $X_1$ never enters the tree structure if $X_2$ is present. By removing the variable $X_2$ from the model, $X_1$ may occur prominently in the splits, leading to a tree as accurate as the previous one. The answer is in the usage of the surrogate splits. Let $\tilde{X}_{j}$ define the surrogate variable, which is a variable that most accurately predicts the action of the best split $s$ on $X_j$ selected by the CART algorithm. The measure of VI of $X_j$ is defined as
\begin{equation} \label{varimp}
VI(X_j)=\sum_{m\in T} \Delta Dev(T)_{ X_j \cup \tilde{X}_j}
\end{equation}
\noindent
where $m$ are the terminal nodes of the tree $T$ and $\Delta Dev(T)_{ X_j \cup \tilde{X}_j}$ represents the decrease of the deviance done by $X_j$ and its surrogate. If there exist more than one surrogate variable for $X_j$ at any node, use the one with larger $\Delta Dev(T)$ in ~\eqref{varimp}. The measure of importance of variables generally used are normalized quantities, $100 VI(X_j)/\max_{j} VI(X_j)$, so the most important feature has measure $100$, the others are in the range from $0$ to $100$. 

For Random Forests, the VI (RF\_VI) is computed by the \emph{Mean Decrease Accuracy} (MDA)\cite{breiman2001b}. Let $X_j$ the variable for which the importance has to be evaluated. The MDA, relies to the permutation principle and involves the OOB estimates. Let $\mathcal{OOB}_{h}$ be the out--of--bag sample of the $h^{th}$ tree. Let $\mathbf{{X}}_{j,\texttt{perm}}$ be the input variable vector where the $j^{th}$ variable has been permuted, and let $\mathcal{OOB}_{h,\texttt{perm} }$ be its corresponding out--of--bag sample. The MDA for $X_j$ is defined by:
\begin{equation}\label{mda}
\begin{split}
{RF\_VI}(X_j)&=  \frac{1}{H} \sum_{h=1}^H  \left[\frac{1}{|\mathcal{OOB}|_{h}} (Y - T_{\texttt{rf}}(\bm {X}))^2\right] -\left[  \frac{1}{|\mathcal{OOB}|_{h,\texttt{perm}}} (Y -
T_{\texttt{rf}}(\bm{X}_{j,\texttt{perm}}))^2 \right] 
\end{split}
\end{equation}
that is the average difference in accuracy of the out--of--bag versus permuted out--of--bag observations over the $H$ trees. This is the variable importance measure for $X_j$. 

For the scope of this paper, we consider two other measures of VI developed in \cite{strobl2008} and \cite{benard2022}: the Conditional VI and the Sobol--MDA VI. The Conditional VI (CF--VI) is defined as:
\begin{equation}\label{cfvi}
\begin{split}
{CF\_VI}(X_j)&=  \frac{1}{H} \sum_{h=1}^H  \left[\frac{1}{|\mathcal{OOB}|_{h}} (Y - T_{\texttt{rf}}(\bm {X}))^2\right] -\left[  \frac{1}{|\mathcal{OOB}|_{h,\texttt{perm}}} (Y -
T_{\texttt{rf}}(\bm{X}_{j,\texttt{perm}|Z}))^2 \right] 
\end{split}
\end{equation}
where $\bm{X}_{j,\texttt{perm}|Z}$ are the values of the variable $X_j$ after permuting its values within the grid defined by the conditioning variables Z.
The rationale for the choice to conditioning also to $Z$ is to decrease the importance of correlated variables, which is generally overestimated in the RF\_VI. To determine the variables Z to be conditioned on, the most conservative choice would be to include all other variables as conditioning variables. Another, more intuitive, choice is to include only those variables whose empirical correlation with the variable of interest exceeds a certain threshold. For the more general case of predictor variables of different scales of
measurement the framework promoted in \cite{hothorn2006} provides p--values of conditional inference tests as measures of association.

The Sobol--MDA VI (S\_MDA-VI) is defined as:
\begin{equation}\label{sobolmda}
\begin{split}
{S\_MDA-VI}(X_j)&= \frac{1}{V(Y)} \frac{1}{H} \sum_{h=1}^H  \left[\frac{1}{|\mathcal{OOB}|_{h}} (Y - T_{\texttt{rf}}(\bm {X}))^2\right]-\\
&\left[  \frac{1}{|\mathcal{OOB}|_{h}} (Y -
T^{\texttt{project}}_{\texttt{rf}}(\bm{X}_{-j}))^2 \right] 
\end{split}
\end{equation}
where $T^{\texttt{project}}_{\texttt{rf}}(\bm{X}_{-j})$ is the projected forest, a random forests obtained as a sum of projected trees: in these kind of trees, the partition of the covariate space obtained with the terminal leaves of the original tree is projected along the $j-th$ direction, and the outputs of the cells of this new projected partition are recomputed with the training data. This allows to obtain the accuracy of the associated out--of--bag projected forest estimate, and by subtracting it from the original accuracy and normalizing by the variance of $Y$ on obtain the Sobol--MDA VI for $X_j$.

\section{Global Sensitivity Analysis}\label{sec:gsa}

In this section, we first review Sobol' sensitivity indices~\cite{sobol1995, Saltelli1995, homma1996}, and then we introduce the GSA paradigm. 

\subsection{Review of sensitivity measures}\label{sec:sm}
 
Sobol' sensitivity indices decompose the variance of the model output $V(Y)$ into terms of increasing dimensionality:
\begin{equation}
V(Y)=\sum_{i=1}^{k}V_i+\sum_{i}\sum_{i<j}V_{ij}+...+V_{1,2,...,k}\,,
\label{eq:decomposition}
\end{equation}
where
\begin{equation}
\begin{aligned}
V_i = V_{X_{i}}\big[E_{\bm{X}_{\sim i}}(Y | X_i)\big] \hspace{4mm} 
V_{ij} &= V_{X_{i}, X_{j}}\big[E_{\bm{X}_{\sim i, j}}(y | X_i, X_j)\big]   \, \\ 
& - V_{X_{i}}\big[E_{\bm{X}_{\sim i}}(Y | X_i)\big] \\
& - V_{X_{j}}\big[E_{\bm{X}_{\sim j}}(Y | X_j)\big]
\end{aligned}
\label{eq:Ex_i}
\end{equation}
and similar relations for the higher order terms~\cite{Saltelli1995}. $V_i$ is that part of the variance $V(Y)$ that is only due to $X_i$, $V_{ij}$ the part due to $X_{i}$ and $X_j$ on $V(Y)$, and so on. $E _{\bm{X}_{\sim i}}(Y|X_i)$ indicates a mean of $Y$ that is taken over all inputs except $X_i$. Sobol' indices are computed as
\begin{equation}
S_i=\frac{V_i}{V(Y)} \hspace{4mm}  S_{ij}=\frac{V_{ij}}{V(Y)} \hspace{4mm} \hdots \,.
\label{eq:Si}
\end{equation}
where $S_i$, $S_{ij}$ are the first order and second order terms due respectively to $X_i$ and  ($X_i,X_j$). $S_i$, $S_{ij}$, ... represent the expected reduction in variance that would be achieve by fixing $X_i$, ($X_i,X_j$) respectively. 

An important measure is the total--order index $T_i$, that includes all terms -- of the first order and higher, that include $X_i$ \cite{homma1996}. $T_i>S_i$, $x_i$ flags the presence of interaction term(s) involving $X_i$ . $T_i$ is defined as  
\begin{equation}
T_i=1 - \frac{V_{\bm{X}_{\sim i}}\big[E_{X_i}(Y | \bm{X}_{\sim i})\big]}{V(Y)} = \frac{E_{\bm{X}_{\sim i}}\big[V_{X_{i}}(Y | \bm{X}_{\sim i})\big]}{V(Y)} \,.
\label{eq:Ti}
\end{equation}

Among the several estimators available to estimate  Equations~\ref{eq:Si}--\ref{eq:Ti} we use those indicated in ~\cite{saltelli2010a}.

\begin{equation}
S_i=\frac{V(Y) - \frac{1}{2N} \sum_{v=1}^{N} \left [f (\bm{B})_v - f(\bm{A}_B^{(i)})_v
 \right ]^2}{V(Y)}\,,
 \label{eq:jansen_si}
\end{equation}

\begin{equation}
T_i= \frac{\frac{1}{2N}\sum_{v=1}^{N} \left [ f(\bm{A})_v - f(\bm{A}_B^{(i)})_v \right ]^2}{V(Y)}\,.
 \label{eq:jansen_ti}
\end{equation}
Further discussion of these estimators is offered in~\cite{saltelli2010a}.  

\subsection{Global Sensitivity Analysis Paradigm}\label{sec:gsa-par}
The sensitivity index reviewed in the previous paragraphs and based on the decomposition of the variance of the model output are the main tools used in GSA. 
To what extent can one talk of a global sensitivity analysis paradigm? It is generally accepted that in model validation on can have two paradigms, one more technical and one more socio--political or participatory ~\cite{Eker_Rovenskaya_Obersteiner_Langan_2018}. The idea has a parallel in sociology of quantification, where one also calls for a double dimension of quality is statistical work ~\cite{Salais_2022,Sen_1990}, again one technical and one normative. A radical vision of sensitivity analysis is offered by ~\cite{Hall_2020} for whom the real strength of the models is in ``sensitivity analysis (where one could examine the response of the model to parameters or structures that were not known with precision." Econometricians  ~\cite{Leamer_1985} and ~\cite{Kennedy_2008} also insist on the next to explore widely around the assumption of a study to see if its inference is robust. More recently these ideas have resurfaced in relation to the attempt to answer the problem of scarce reproducibility of quantitative analysis ~\cite{Gelman_Loken_2013}, with actual experiments performed by several teams attempting to replicate the same analysis ~\cite{Breznau_Rinke_Wuttke_2022} -- a line of investigation that has revealed ``A universe of uncertainty hiding in plain sight" ~\cite{Engzell_2023}. Adopting sensitivity analysis -- or better GSA ~\cite{Saltelli_Tarantola_Campolongo_2000} as a paradigm, implies tackling this uncertainty head on, by propagating all plausible uncertainties and ambiguities -- be these technical or normative, through the assessment. Some investigators call this ``multiverse analysis" ~\cite{Steegen_Tuerlinckx_Gelman_Vanpaemel_2016}, while we call it either global sensitivity analysis or in a more verbose definition ``modelling of the modelling process" ~\cite{Saltelli_Puy_2023}.     

\section{Global Sensitivity--Based Ranking of Variables for Explainable Random Forests}\label{sec:GS-VI}

To enhance the explainability of RF VI, we propose to use the total sensitivity measure of Equation \ref{eq:jansen_ti} as an indicator of VI measure in a GSA paradigm: the RF\_GS-VI. The total sensitivity index in GSA is based on a systematic exploration of the space of the model input to get the influence on the model output. We estimate $S_{Ti}$ using an estimator and a structured sample constructed as in \cite{saltelli2010a}. This kind of estimator is employed also in \cite{becker2021} to rank regressors in terms of their importance in a regression model. Specifically, given the sample data $ \left \{Y_{i},\bm{X}_{i}\right \} $ let us consider $\bm{\gamma}$ RF models. Let $q(\bm{\gamma})$ be a measure of the model fit. Since the BIC and AIC measures used  by \cite{becker2021} are not applicable to models as RF, we proposed as $q(\bm{\gamma})$ the Root Mean Squared Error (RMSE). In order to compute the $S_{Ti}$ one should be able to compute all $q(\bm{\gamma})$ for all $\bm{\gamma} \in \bm{\Gamma}$, which could be practically unfeasible. Therefore, generate a random draw of $\bm \gamma$ in $\bm \Gamma$, say $\gamma_*$; then consider elements $\gamma_*^{(i)}$ with all elements equal to $\gamma_*$ except for the i--th coordinate which is switched from 0 to 1 or vice--versa. This is used to calculate the $VI(\bm \gamma)$ and apply the estimator of \cite{saltelli2010a}. This process is described in Algorithm \ref{myalgo}.

	\begin{algorithm}
		\footnotesize
		$ \left \{Y_{i},\bm{X}_{i}\right \} $, $i=1,\dots,n$, $\text{length}(\bm{X}_{i})=p$ \\
		 \For{$l=1$ \KwTo $L$}{
    Sample $\gamma_l$ in $\bm \Gamma$ $\sim \textrm{Discrete Uniform}(0,1)$\;
    Fit $Y= T_{\texttt{rf}}({X_{\gamma_l}})$\;
    Evaluate $q_l$ = $q(\gamma_l)$\;
    \Repeat{$k=p$}{
        Take the $k-th$ element of $\gamma_l$, and switch it to 0 if it is equal to 1, and to 1 if it is 0.\\ Denote this new vector with inverted  $k-th$ element as $\gamma_l^{(k)}$\;
    Evaluate $q_{kl}$ = $q(\gamma_l^{(k)})$\;
    }
    }
		$ \hat{S}_{T_i} = \frac{\hat{\sigma}^2_{Ti}}{\hat{V}}=\frac{\hat{\sigma}^2_{Ti} = \frac{1}{4N} \sum_{l=1}^{N} (q_{kl}-q_l)^2}{\hat{V} = \frac{1}{N-1} \sum_{l=1}^{N} (q_l- \bar{q})^2} $.
		\caption{Pseudocode for RF\_GS-VI}\label{myalgo}
	\end{algorithm}

\section{Simulation study}\label{sec:sim}

The simulation study is aimed to explore some data generating process (DGP) in terms of generative importance between predictors and response variable.
We explored three scenarios that cover different type of regression functions.

\textsc{Scenario 1}
	\begin{equation}\label{eq1}
		\begin{split}
            X_1 &= \epsilon_1\\
			X_j &=a X_1+\epsilon_j \qquad\text{for}\; j=2,\dots,4 \\
			Y&=b X_2+b X_3+b X_4+\epsilon_Y.
		\end{split}
	\end{equation}
where the response $Y$ depends directly on three intermediate variables $X_2$, $X_3$ and $X_4$, and it is only indirectly dependent on the background variable $X_1$. Therefore,  $Y \ind X_1 | X_2, X_3, X_4$. We set $a=b=3$, $\epsilon_j \sim N(0,1)$, $j=1, \ldots 4$, $\epsilon_Y \sim N(0,1)$, $Y=1, \ldots n$. 

\textsc{Scenario 2} 
	\begin{equation}\label{eq2}
		\begin{split}
            X_1 &= \epsilon_1\\
            X_2 &= a X_1+\epsilon_2\\
			X_j &=a X_2+\epsilon_j \qquad\text{for}\; j=3,4 \\
            X_5 &=a X_2+\epsilon_5\\
			Y&=b X_3+b X_4+\epsilon_Y.
		\end{split}
	\end{equation}
where the response $Y$ depends directly on two intermediate variables $X_3$ and $X_4$, and it is only indirectly dependent on the background variables $X_1$ and $X_2$. Moreover, there is no direct association between $Y$ and $X_5$. Therefore, $ Y \ind X_1| X_2$, $ Y \ind X_2| X_3 , X_4$, $Y \ind X_5 | X_2$ and
$Y \ind X_5 | X_3, X_4$. We set $a=b=3$, $\epsilon_j \sim N(0,1)$, $j=1, \ldots 5$, $\epsilon_Y \sim N(0,1)$, $Y=1, \ldots n$.

 \textsc{Scenario 3} 
	\begin{equation}\label{eq3}
		\begin{split}
            X_1 &= \epsilon_1\\
            X_2 &= a X_1 + \epsilon_2\\
			X_3 &= \epsilon_3 \\
			Y&=b X_2+b X_3+\epsilon_Y \\
            X_4 &= c X_2 + c Y+\epsilon_4\\
            \end{split}
	\end{equation}
 
 where the response $Y$ depends directly on two intermediate variables $X_2$ and $X_3$, and it is only indirectly dependent on the background variable $X_1$. Moreover, the covariate $X_4$ is directly dependent to the response and the intermediate variable $X-2$. Therefore, $  Y \ind X_1| X_2$. We set $a=b=3$ and $c=2$, $\epsilon_j \sim N(0,1)$, $j=1, \ldots 4$, $\epsilon_Y \sim N(0,1)$, $Y=1, \ldots n$. This scenario represents an interesting case in which there is a variable affected by the response, but the researcher includes all variables as explanatory variables.
 
\begin{figure}%
    \centering
   \includegraphics[scale = 0.95]{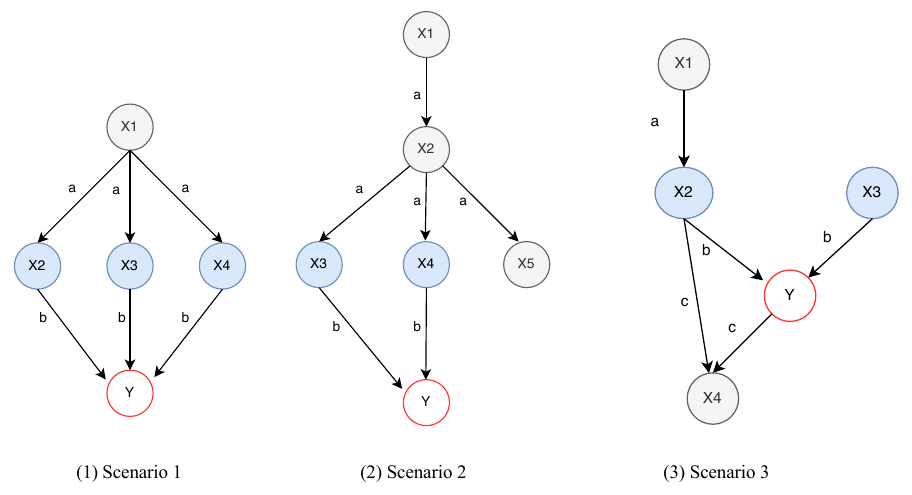}
    \caption{The Direct Acyclic Graph (DAG) of the recursive regression systems for equations \ref{eq1} (1), \ref{eq2} (2) and \ref{eq3} (3).
}%
    \label{fig:dag}%
\end{figure}
 
Figure \ref{fig:dag} shows the Direct Acyclic Graph (DGP) for each system of regression equations depicted in the three scenario, while Table \ref{tab:condind} summarizes the resulting conditional independences between the response and the explanatory variables and among the explanatory variables.

\begin{table}[H]
	\renewcommand*{\arraystretch}{1.2}
	\setlength{\tabcolsep}{3pt}
	\label{tab:condind}
	\centering
	\begin{sc}
\begin{tabular}{|l|c|}
\hline
           &    Conditional Independences \\ \hline
Scenario 1 &    \multicolumn{1}{c|}{\begin{tabular}[c]{@{}c@{}}
$ X_2 \ind X_3 | X_1$;
$ X_2 \ind X_4 | X_1$;
$ X_3 \ind X_4 | X_1$;
\\ $Y \ind X_1 | X_2, X_3, X_4$
\end{tabular}}                      
\\ \hline
Scenario 2  &  \multicolumn{1}{c|}{\begin{tabular}[c]{@{}c@{}}
$ X_1 \ind X_3 | X_2$;
$ X_1 \ind X_4 | X_2$;
$ X_1 \ind Y| X_3 , X_4$;
$ X_1 \ind X_5 | X_2$;\\
$ X_3 \ind X_4 | X_2$;
$ X_3 \ind X_5 | X_2$;
$ X_4 \ind X_5 | X_2$;\\
$ Y \ind X_1| X_2$;
$ Y \ind X_2| X_3 , X_4$;
$Y \ind X_5 | X_2$;
$Y \ind X_5 | X_3, X_4$
\end{tabular}}                       \\ \hline
Scenario 3 & \multicolumn{1}{c|}{\begin{tabular}[c]{@{}c@{}}
$ X_1 \ind X_3$;
$ X_1 \ind X_4 | X2$;
$ X_2 \ind X_3 $; 
$ X_3 \ind X_4 | X_2, Y$; 
$  Y \ind X_1| X_2$
\end{tabular}}             
\\ \hline
		\end{tabular}
	\end{sc}
 \caption{Conditional independences between the response and the explanatory variables and among the explanatory variables for the three DGP proposed.}
\end{table}

All the results of this paper were obtained by \texttt{R} statistical software (R Core Team 2023, version 4.2.3). For evaluation purposes, we considered a sample size of $n=1000$ for each scenario, and we investigated $1000$ data sets. We compared:
\begin{itemize}
\item \textbf{RF\_GS-VI}, our proposal, implemented as a user--written function with the usage of the function \texttt{randomForest};
\item \textbf{CART-VI}, the VI measure of CART algorithm proposed by \cite{breiman1984} and implemented in the package \texttt{rpart}, function \texttt{rpart};
\item \textbf{RF-VI}, the VI measure of RF algorithm proposed by \cite{breiman2001a} and implemented in the package \texttt{randomForest}, function \texttt{randomForest};
\item \textbf{CF-VI}, the Conditional VI proposed by \cite{strobl2008} and implemented in the package \texttt{party}, functions \texttt{cforest} and \texttt{varimp};
\item \textbf{S\_MDA-VI}, the VI measure for RF proposed by \cite{benard2022} and implemented in the package \texttt{SobolMDA} based on the fast RF package \texttt{ranger}, function \texttt{ranger}.

\end{itemize}

\begin{figure}
		\centering
		\includegraphics[scale = 0.37]{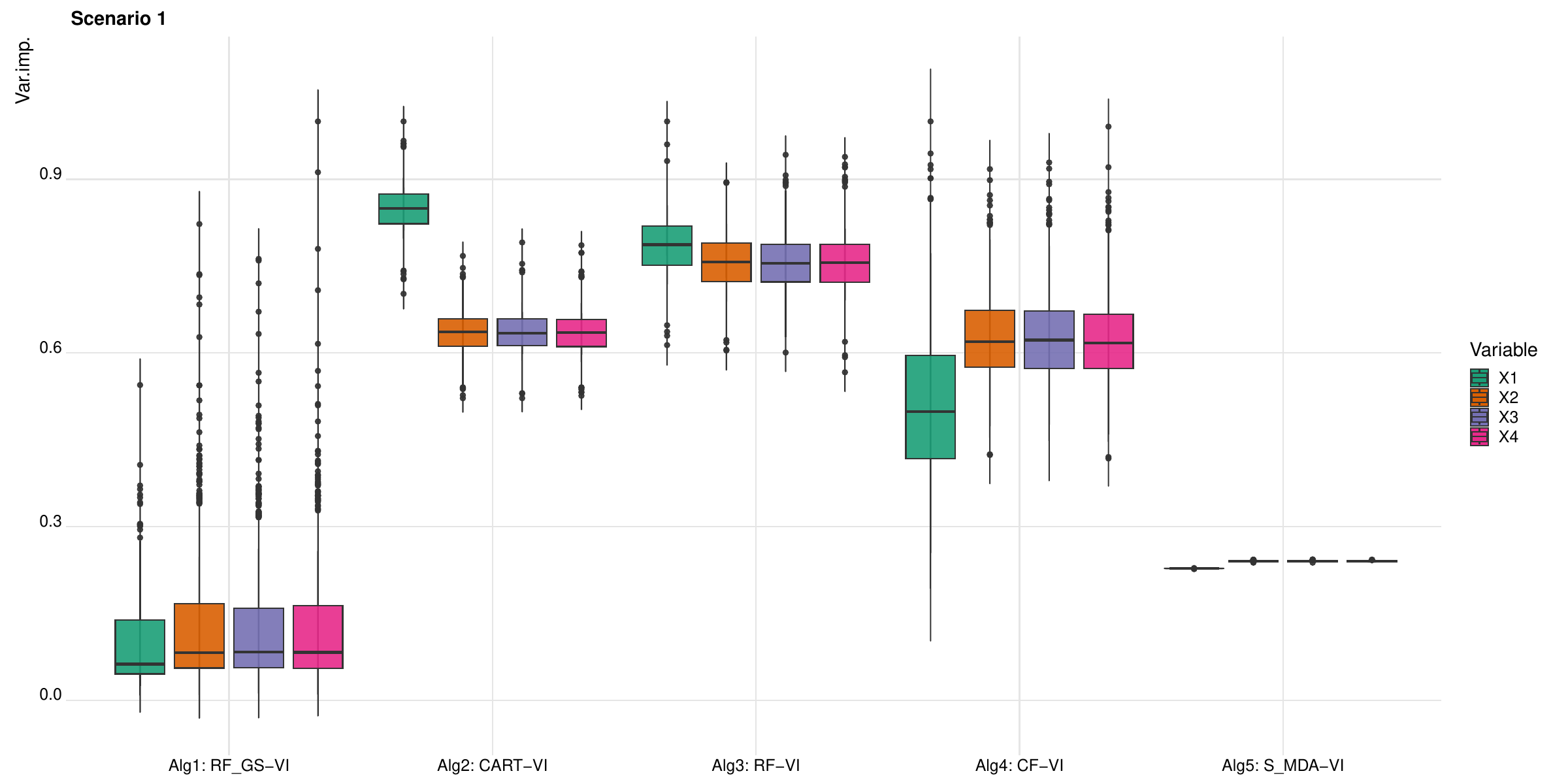}
		\caption{Violin plot of Monte Carlo distributions of RF\_GS-VI, CART-VI, RF-VI, CF-VI and S\_MDA-VI for the Scenario 1, $n = 1000$.}
  \label{fig:sc1}
	\end{figure}

\begin{figure}
		\centering
		\includegraphics[scale = 0.37]{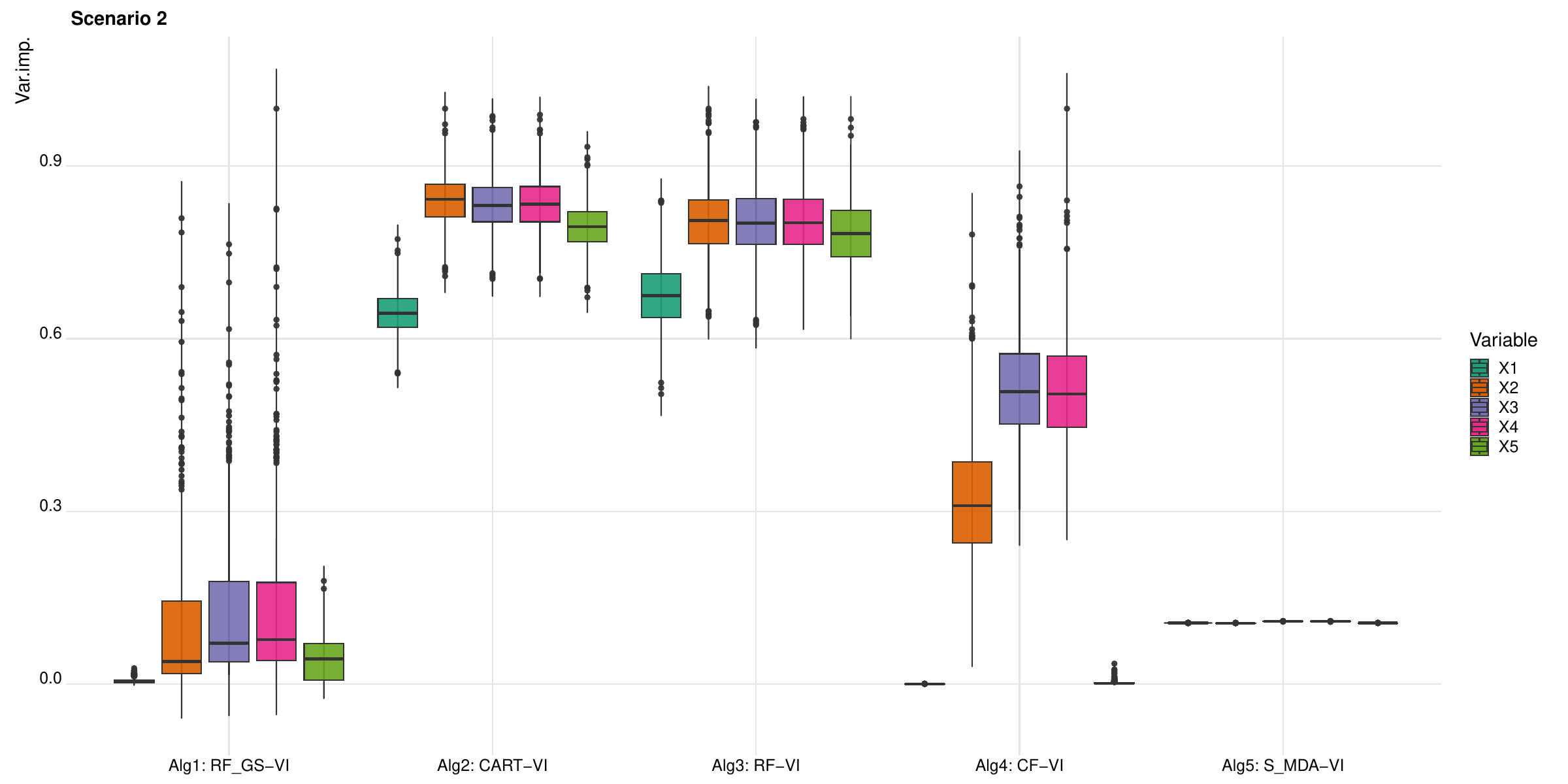}
		\caption{Violin plot of Monte Carlo distributions of RF\_GS-VI, CART-VI, RF-VI, CF-VI and S\_MDA-VI for the Scenario 2, $n = 1000$.}
  \label{fig:sc2}
	\end{figure}

\begin{figure}
		\centering
		\includegraphics[scale = 0.37]{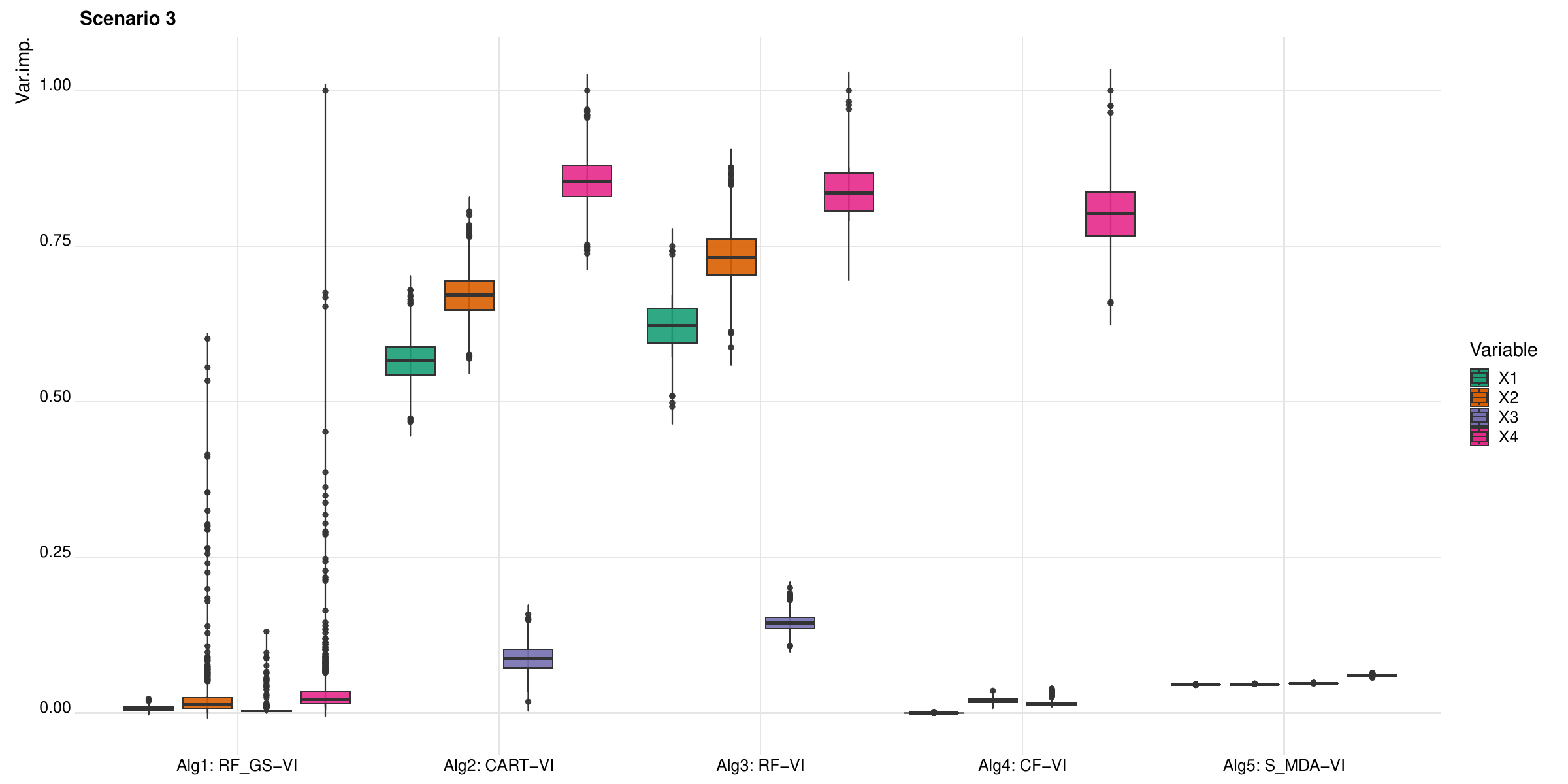}
		\caption{Violin plot of Monte Carlo distributions of RF\_GS-VI, CART-VI, RF-VI, CF-VI and S\_MDA-VI for the Scenario 3, $n = 1000$.}
   \label{fig:sc3}
	\end{figure}

In this simulation study we have examined the Monte Carlo distributions of the above VI: the violin plots (boxplot and density plot) are shown in Figure \ref{fig:sc1} for the DGP of scenario 1, Figure \ref{fig:sc2} for the DGP of scenario 2 and Figure \ref{fig:sc3} for the DGP of scenario 3, while Table \ref{VI-table2} summarises these distributions in terms of Monte Carlo means and standard deviations (in brackets). Table \ref{prop_table} also shows the proportions of variables correctly ranked in the first position. According to scenario 1, we considered a correct ranking in the first position if the variable is $X_2$, $X_3$ or $X_4$. For scenario 2, we considered a correct ranking at the first position if the variable is $X_3$ or $X_4$. For scenario 3, we considered a correct ranking at the first position if the variable is $X_2$ or $X_3$.

For scenario 1, figure \ref{fig:sc1} (and table \ref{VI-table2}) shows that with the proposed RF\_GS-VI, the mean of the distribution of VIs is lower for the variable $X_1$ and higher for the variables $X_2$, $X_3$ and $X_4$. Also for CF-VI and S\_MDA-VI we obtain a lower mean of the VI for the variable $X_1$ with respect to the other variables. On the contrary, for both CART-VI and RF-VI, the highest value of the mean is obtained from $X_1$. From the proportions of correct variables in the first position of the ranking of the VIs in Table \ref{prop_table}, it can be seen that with our proposal, approximately $80\%$ of the Monte Carlo replications have a correct variable in the first position of the ranking. For RF-VI, this percentage decreases to about $60\%$ of the cases, while for CART-VI, $X_1$ is the most important variable in almost all Monte Carlo replications. For CF-VI, about $88\%$ of the Monte Carlo replications have a correct variable in the first position of the ranking, while in the total of the Monte Carlo replications with S\_MDA-VI, a correct variable is obtained in the first position. 

For scenario 2, figure \ref{fig:sc2} (and table \ref{VI-table2}) shows that with the proposed RF\_GS-VI, the mean of the distribution of the VIs is lower for $X_1$ and $X_5$, and it is higher for the variables $X_2$, $X_3$ and $X_4$, with the maximum values of the mean reached by the VIs of $X_3$ and $X_4$. For the CART-VI and RF-VI means, the minimum value is obtained from $X_1$ and $X_5$, while for all the other variables the means are quite similar. For CF-VI, the highest values of the means are obtained for the variables $X_3$ and $X_4$, and then smaller values for the variables $X_2$, followed by $X_5$ and $X_1$. The same results are obtained from S\_MDA-VI. From the proportions of correct variables in the first position of the ranking of the VIs in Table \ref{prop_table}, it can be seen that with our proposal, about $70\%$ of the Monte Carlo replications have a correct variable in the first position of the ranking. This percentage decreases to about $56\%$ of the cases for RF-VI, while for CART-VI it is about $70\%$. For CF-VI and for the sum of the Monte Carlo replications, a correct variable is obtained in the first position. 

For scenario 3, figure \ref{fig:sc3} (and table \ref{VI-table2}) shows that with our proposal, the mean of the distribution of VIs is lower for $X_1$ and $X_3$, and it is higher for the variables $X_2$ and $X_4$, with the VI of $X_4$ reaching the maximum value. From the averages of CART-VI, RF-VI, CF-VI and S\_MDA-VI, the minimum value is reached by $X_3$, while for all the other variables the averages are quite similar, with the VI of $X_4$ always reaching the maximum value. From the proportions of correct variables in the first position of the ranking of the VIs in the table\ref{prop_table}, it can be seen that with our proposal about $26\%$ of the Monte Carlo replications have a correct variable in the first position of the ranking. For all other algorithms considered, this percentage is about $0\%$ of the cases.

 \begin{table}
	\renewcommand*{\arraystretch}{1.2}
	\setlength{\tabcolsep}{3pt}
	\label{VI-table2}
	\centering
	\begin{sc}
		\small		  
		\begin{tabular}{lrrrrr} \hline
			& $X_1$ & $X_2$ & $X_3$ &  $X_4$ & $X_5$ \\
			\hline   
   \hline
   Scenario 1 \\
   \hline
   			RF\_GS-VI    & 1.8 &2.3 &2.3 &2.3 \\
   & (1.2) & (1.9) & (1.9) &
    (1.9 )\\
			CART-VI      & 705\,035.2 & 527\,604.2 & 527\,526.7  & 
   527\,820.5\\ 
			& (33\,004.9) & (30\,279.9) & (29\,849.3) &
   (30\,674.84)\\
			RF-VI  & 194\,304.5 & 186\,668.7  & 186\,681.3 &
   186\,659.2\\
			& (12\,615.4) & (12\,203.9) & (12\,279.0) &
    (12\,488.8)\\
    	CF-VI  & 8.0 &9.8 &9.9 &9.8 \\
			& (2.1) & (1.2) & (1.2) &
    (1.2)\\
    	S\_MDA-VI  & -0.002 & 0.011  &0.011 & 0.011 \\
			& (0) & (0.001) & (0.001) &
    (0.001)\\
	\hline    
   Scenario 2 \\
   \hline
   			RF\_GS-VI    & 0.1 &1.6 &2.0 &2.1 &0.7 \\
   & (0.06) & (1.8) & (1.9) &
    (2.0) &(0.6)\\
				CART-VI      & 2\,286\,063 & 2\,984\,104 & 2\,957\,274  & 2\,959\,452 & 2\,820\,692 \\ 
			& (128\,474.4) & (148\,527.5) & (156\,767.9) &
   (159\,303.7) &(137\,696.6)\\
			RF-VI  & 567\,700.4 & 675\,789.5   & 674\,645.8  &
   675\,950.3 & 658\,084.8 \\
			& (46\,294.1) & (48\,511.0) & (49\,407.6) &
    (47\,864.7) &(48\,150.01)\\
   CF-VI  & 0 &27.6 &44.3 &43.9&0.1 \\
			& (0.004) & (9.191) & (7.935) &
    (7.717)&(0.200)\\
    	S\_MDA-VI  & 0 & 0 &0.003 & 0.003&0 \\
			& (0) & (0) & (0) &
    (0)&
    (0)\\
			\hline  
   Scenario 3 \\
   \hline
   			RF\_GS-VI    & 1.1 &3.7 &0.8 &5.4 \\
   & (0.7) & (7.6) & (1.5) &
    (9.2)\\
				CART-VI     & 66\,667.6 & 75\,420.7 & 9\,841.8  & 
   96\,045.5\\ 
			& (3\,845.5) & (4\,024.7) & (2\,452.3) &
   (4\,273.3)\\
			RF-VI  & 26\,470.3 & 31\,174.3  & 6\,196.0 &
   35\,655.0\\
			& (1\,764.4) & (1\,825.5 ) & (600.6) &
    (1\,865.1)\\
CF-VI  &0 &0.5 &0.4 &19.3 \\
			& (0.005) & ( 0.093) & (0.177) &
    (1.222)\\
    S\_MDA-VI  & 0 & 0 &0.002 & 0.014 \\
			& (0) & (0) & (0) &
    (0.001)\\
      \hline  
		\end{tabular}
	\end{sc}
 \caption{Monte Carlo averages and standard deviations (in parentheses) of VI measures for simulated data generated according to the DAG of Figure~\ref{fig:dag}.}
\end{table}

 From the simulation study it is clear that it is very easy to get a wrong ranking of variables when using the classical VI measures of CART and RF to infer the underlying generating process of the data. Scenario 1 represents a case also studied in \cite{gottard2020}, where the role of induced correlations in greedy search algorithms such as CART and RF is highlighted. In this case, using the GSA algorithm with the RF algorithm can help to restore the correct ranking of variables in the generative sense. Note that even using VI from Conditional VI for Random Forests and Sobol-MDA VI the correct ranking is obtained. Looking at the proportions of correct variables in the first position of the ranking of VI in this simulation study, the proposed algorithm ranks one of the correct variables in the first position almost $80\%$ of the time, while RF-VI gives a correct variable in the first position in about half of the simulations and CART-VI is practically always wrong. Scenario 2 represents a more difficult DGP, where the VIs of CART and RF on average do not discriminate well between direct and indirect relations between the variables $X_2$, $X_3$ and $X_4$, while the proposed algorithm gives a correct sorting on average. The Conditional VI for Random Forests and the Sobol-MDA VI also give correct rankings. Looking at the proportions of correct variables at the first position of the VI ranking, the proposed algorithm and the CART algorithm give the same proportion, while there is a gain in using our proposal with respect to the RF algorithm. Interestingly, with CF-VI and S\_MDA-VI, we obtain a correct variable in the first position in all the simulations. Finally, scenario 3 is a very interesting case study also in terms of causal inference, and while the VIs of CART, RF, Conditional VI for RF and Sobol-MDA VI always put $X_4$ as the most important variable, our method manages to return a correct variable in the first position in at least $30\%$ of the cases.

In this simulation study, the resulting VI distributions from our proposal show greater variability than those obtained with the other algorithms. This may be helpful to the researcher, since using our method alongside the VI of more well--known methods such as CART and RF may shed light on a discrepancy between the ordering of the variables, which could be due precisely to a different predictive and generative ranking. Finally, it should be noted that the Sobol--MDA proposal achieves excellent performance for scenarios 1 and 2, but when looking at the values of the VIs, they are practically all similar and close to 0. This may lead to erroneous conclusions for the researcher less familiar with the theory underlying the methodology.

 \begin{table}
	\renewcommand*{\arraystretch}{1.2}
	\setlength{\tabcolsep}{3pt}
	\label{prop_table}
	\centering
	\begin{sc}
		\small		  
		\begin{tabular}{lrrrrr} \hline
			& RF\_GS-VI & RF-VI & CART-VI & CF-VI & S\_MDA-VI \\
			\hline   
   \hline
			Scenario 1      & 0.794 & 0.526 & 0.001 &0.875&1 \\
			Scenario 2  & 0.696 & 0.564  & 0.698 &1&1\\
			Scenario 3    & 0.264 &0.018 &0 &0&0 \\
	\hline    

		\end{tabular}
	\end{sc}
 	\caption{Proportions of correct variable at first position of ranking of VI for each scenario and algorithm.}
\end{table}

\section{Applications}\label{sec:appl}
In this section we compute the VI for the five approaches, RF\_GS-VI, RF-VI, CART-VI, CF-VI and S\_MDA-VI for two data sets: the energy efficiency data set \cite{misc_energy_efficiency_242}, and the liver disorders data set \cite{misc_liver_disorders_60}, both available on the UC Irvine Machine Learning Repository (https://archive.ics.uci.edu). The energy efficiency study concerns energy analysis using 12 different building shapes simulated in Ecotect. The buildings differ in terms of glazing area, glazing area distribution, orientation, height, wall area, roof area, floor area and relative compactness. After simulating various settings as a function of the above characteristics, 768 building shapes were obtained. Therefore, the data set consists of $n=768$ samples and $X_p=8$ explanatory variables, with the aim of predicting two real--valued responses, the heating load ($Y_1$) and the cooling load ($Y_2$). The study on liver diseases concerns liver diseases caused by excessive alcohol consumption.  The data set consists of $n=345$ male individual and $X_p=5$ explanatory variables, which represents blood parameters that are thought to be sensitive to liver disorders. Specifically, these blood parameters are the mean corpuscular volume, the alkaline phosphotase, the alanine aminotransferase, the aspartate aminotransferase and the gamma-glutamyl transpeptidase. The task is to predict a real--valued response, the number of half--pint equivalents of alcoholic beverages drunk per day $Y$.

We perform a $5-$fold CV on the dataset and compute for our proposal, RF, Conditional VI for RF and Sobol-MDA $100$ Monte Carlo replications within the folds to obtain the VI. We then average the values of the VI across the folds and obtain the Monte Carlo distributions of the VI relatives to each explanatory variable. For the CART algorithm, we obtain the VI for each explanatory variable only from the $5-$ folds and average the values across these folds. 

\begin{figure}
		\centering
		\includegraphics[scale = 0.45]{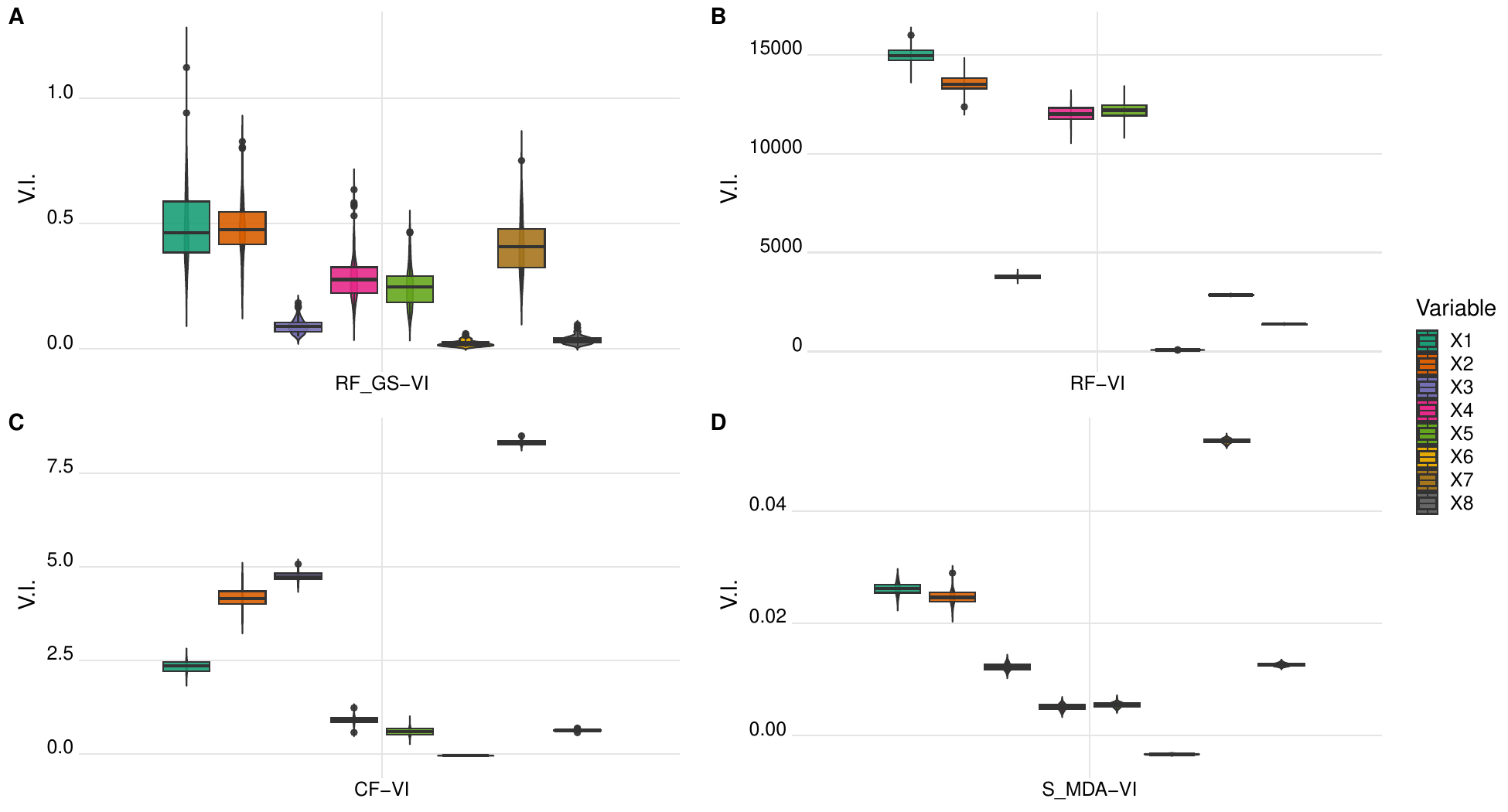}
		\caption{Energy data set analysis: boxplots of $100$ Monte Carlo replications of VI measures for $Y_1$: in A of RF\_GS-VI; in B of RF-VI; in C of CF-VI; in D of S\_MDA-VI.}
  \label{fig:resy1}
	\end{figure}

\begin{figure}
		\centering
		\includegraphics[scale = 0.45]{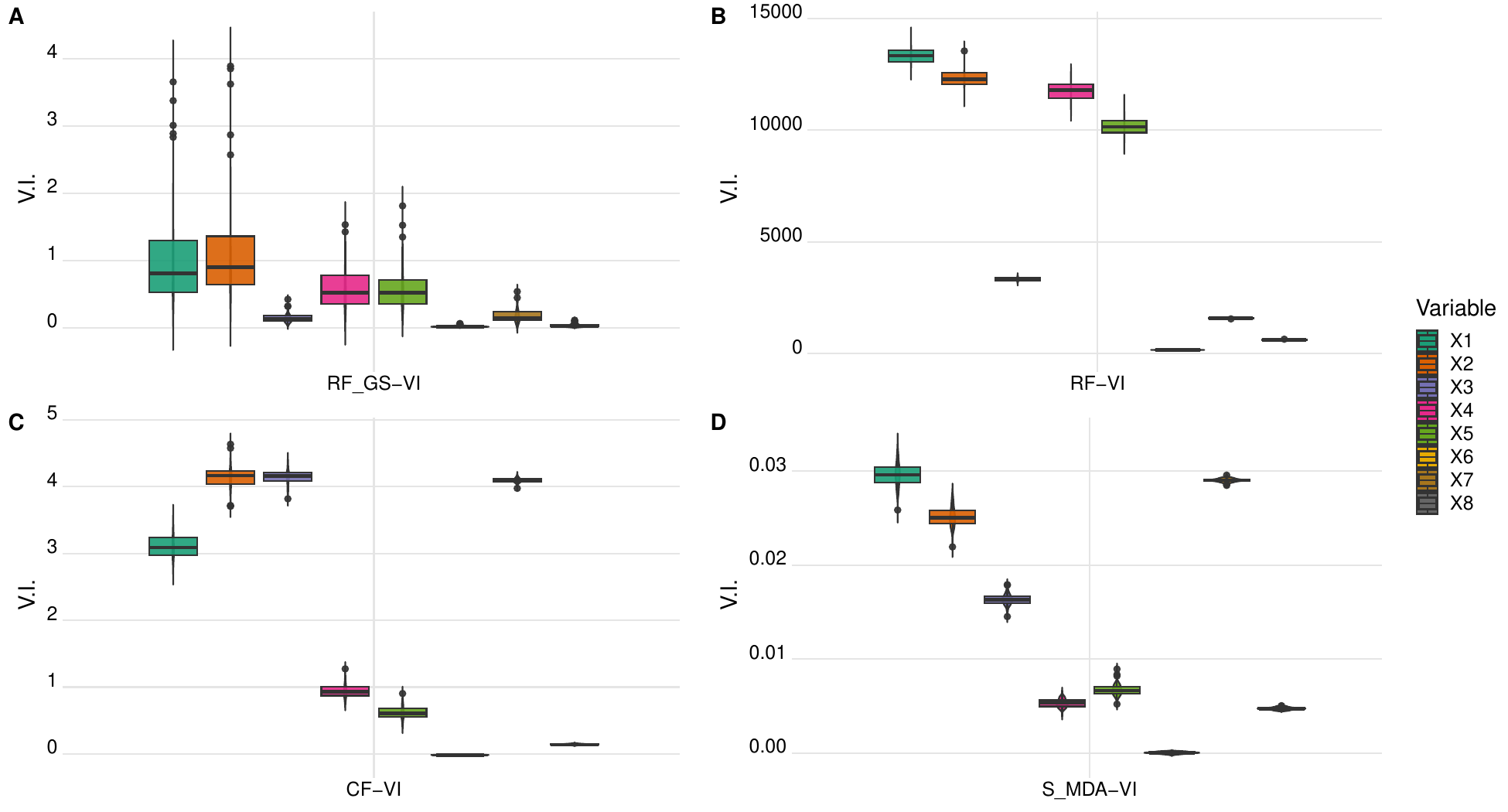}
		\caption{Energy data set analysis: boxplots of $100$ Monte Carlo replications for $Y_2$: in A of RF\_GS-VI measure; in B of RF-VI measure; in C of CF-VI measure; in D of S\_MDA-VI measure.}
  \label{fig:resy2}
	\end{figure}

 For the energy efficiency dataset, the plots of the Monte Carlo replications for RF\_GS-VI, RF-VI, CF-VI and S\_MDA-VI are shown in Figures \ref{fig:resy1} and \ref{fig:resy2} for the models for $Y_1$ and $Y_2$ respectively. For CART-VI, the average of VI over the 5-folds is given in table \ref{cart-vi-energy}. Note that for CART-VI the VI for $X_6$ is not reported as it is not calculated by the algorithm in the majority of folds. For $Y_1$, RF\_GSA-VI reports $X_1$, $X_2$, $X_7$, $X_4$ and $X_5$ as the variables with the highest values of VI, both RF-VI and CART-VI report $X_1$, $X_2$, $X_4$ and $X_5$ as the variables with the highest values of VI, while both CF-VI and S\_MDA-VI indicate $X_7$, $X_3$ and $X_2$ as the variables with the highest values of VI. In this case we can conclude that our proposal can highlight the variable $X_7$ as a possible influential variable, and this is also confirmed by the conditional VI for RF and Sobol-MDA, while with the CART and RF VIs this variable would never be considered as influential. For $Y_2$, RF\_GSA-VI, RF-VI and CART-VI give $X_1$, $X_2$, $X_4$ and $X_5$ as the variables with the highest values of VI, while CF-VI gives $X_2$, $X_3$ and $X_7$ as the variables with the highest values of VI, and S\_MDA-VI gives $X_1$, $X_2$ and $X_7$ as the variables with the highest values of VI.  Therefore, for $Y_2$ we can conclude that our proposal confirms the ranking of the RF and CART algorithms, while the conditional VI for RF and Sobol--MDA also highlights the importance of $X_7$, while it seems to be able to consider $X_4$ and $X_5$ as less influential.

 \begin{table}
	\renewcommand*{\arraystretch}{1.2}
	\setlength{\tabcolsep}{3pt}
	\label{cart-vi-energy}
	\centering
	\begin{sc}
		\small	
\begin{tabular}{lll}
\hline 
                                                  & \multicolumn{2}{c}{\hspace{20mm}CART-VI}                           \\ \hline \hline
\multicolumn{1}{c}{} & \multicolumn{1}{c}{\hspace{20mm}$Y_1$} & \multicolumn{1}{c}{\hspace{20mm}$Y_2$} \\ \hline
$X_1$                                    &\hspace{20mm}54\,699.203&\hspace{20mm}48\,750.940\\ \hline
$X_2$                                    &\hspace{20mm}54\,699.203&\hspace{20mm}48\,750.940\\ \hline
$X_3$                                    &\hspace{20mm}21\,666.449&\hspace{20mm}18\,965.702\\ \hline
$X_4$                                    &\hspace{20mm}49\,429.970&\hspace{20mm}44\,561.670\\ \hline
$X_5$                                    &\hspace{20mm}49\,429.970&\hspace{20mm}44\,561.670\\ \hline
$X_6$                                    &\hspace{20mm}-&\hspace{20mm}-\\ \hline
$X_7$                                    & \hspace{22mm}4\,052.775& \hspace{22mm}1\,164.655\\ \hline
$X_8$                                    & \hspace{22mm}1\,730.266&  \hspace{24mm}508.153\\ \hline
\end{tabular}
\end{sc}
\caption{Energy data set analysis: average across the 5--fold CV of CART-VI for each explanatory variable and the two responses.}
\end{table}

\begin{figure}
		\centering
		\includegraphics[scale = 0.45]{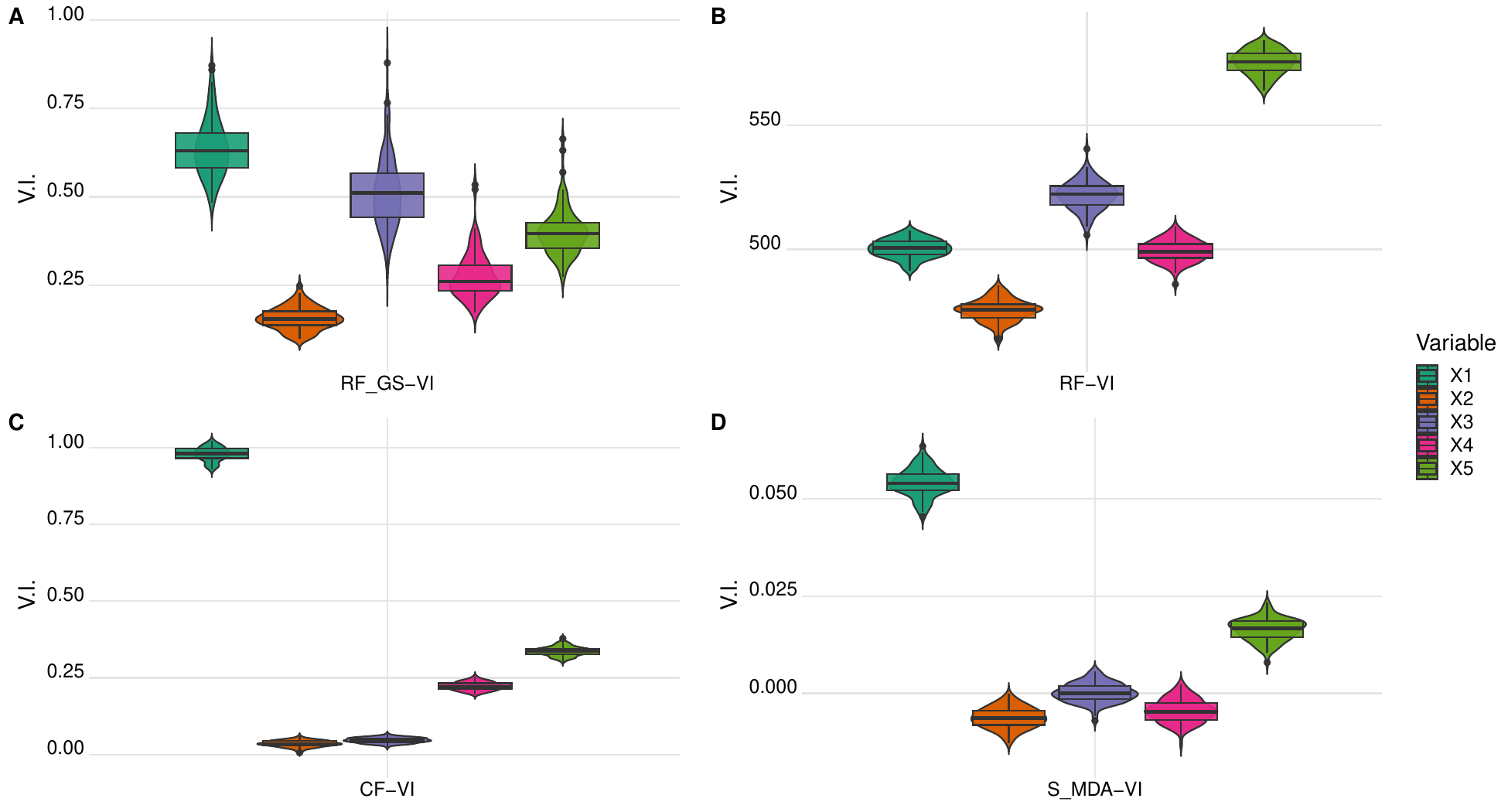}
		\caption{Liver disorders data set analysis: boxplots of $100$ Monte Carlo replications for $Y$: in A of RF\_GS-VI measure; in B of RF-VI measure; in C of CF-VI measure; in D of S\_MDA-VI measure.}
  \label{fig:resbupa}
	\end{figure}

For the Liver disorders data set, the plots of the Monte Carlo replications for RF\_GS-VI, RF-VI, CF-VI and S\_MDA-VI are shown in Figure \ref{fig:resbupa}. For CART-VI, the average VI over the 5 replicates is shown in Table \ref{cart-vi-bupa}. From Figure \ref{fig:resbupa}, RF\_GSA-VI indicates $X_1$ and $X_5$ as the variables with the highest values of VI, RF-VI indicates $X_5$ and $X_3$ as the variables with the highest values of VI, while both CF-VI and S\_MDA-VI indicate $X_1$ and $X_5$ as the variables with the highest values of VI. From the table \ref{cart-vi-bupa}, CART-VI shows $X_4$ and $X_5$ as the variables with the highest values of VI. In this case, we can conclude that all the tested algorithms highlight the importance of the variable $X_5$, while our proposal can highlight the variable $X_1$ as a possible influential variable, and this is also confirmed by the conditional VI for RF and Sobol--MDA, while with the VIs of CART and RF this variable would not be considered as influential.

  \begin{table}[H]
	\renewcommand*{\arraystretch}{1.2}
	\setlength{\tabcolsep}{3pt}
	\label{cart-vi-bupa}
	\centering
	\begin{sc}
		\small	
\begin{tabular}{lc}
\hline 
                                                  & \multicolumn{1}{c}{\hspace{20mm}CART-VI}                           \\ \hline \hline
$X_1$                                    & \hspace{20mm}355.670               \\ \hline
$X_2$                                    & \hspace{20mm}222.046                 \\ \hline
$X_3$                                    & \hspace{20mm}382.772                  \\ \hline
$X_4$                                    & \hspace{20mm}465.039             \\ \hline
$X_5$                                    & \hspace{20mm}442.684                      \\ \hline
\end{tabular}
\end{sc}
	\caption{Liver disorders data set analysis: average across the 5-fold CV for CART-VI for each explanatory variable and the two responses of Liver disorders data set.}
\end{table}

	\section{Conclusions}\label{sec:concl}

Machine Learning techniques are increasingly used in a number of scientific domains, by a population of users at present largely exceeding the restrict community of data analysis practitioners. As a result, the Variable Importance feature of supervised algorithms may happen to be interpreted in terms of generative importance rather than in the more precise and circumscribed terms of predictive importance. In this paper we apply Global Sensitivity Analysis to Random Forests to rank the input features by their generative importance, with the expectation that this will contribute to the explainability of the machine learning methods in general.

From a theoretical point of view, our algorithm can be extended to all kinds of supervised learning algorithms with VI features, i.e. to boosting trees. Furthermore, the GSA paradigm presented here can be extended to other statistical models. \cite{becker2021} applied this paradigm to variable selection in regression. From the simulation study and the applications, our results seem to indicate global sensitivity analysis in indeed effective in elucidating the data generating process for the example where the it is known, and that the case can also be made for the application where it is not. More specifically, global sensitivity analysis can only fail in the case scenario 3, where the response $Y$ depends directly on two intermediate variables $X_2$ and $X_3$, and it is only indirectly dependent on the background variable $X_1$. This represents a challenging situation for causal inference, for which currently none of the algorithms used in this article manages to detect the qualitative difference between $X_3$ and $X_1$, and if the latter is accidentally inserted among the predictors, all the tree--based algorithms tested choose it as the most important variable. On the contrary, our proposal succeeds in this task about $30\%$ of the time.

Further work is needed to determine whether the use of a measure such as the random mean squared error RMSE used in our proposal can be improved upon. Also to be addressed are the theoretical properties of our measure, and a theory--based measure of its error.  

\bibliographystyle{unsrt}  
\bibliography{bibtree}

\end{document}